%% file: main.tex
\documentclass[sigconf]{acmart}
\AtBeginDocument{%
  }

\acmConference[Preprint]{}
\acmBooktitle{In submission.}

\usepackage{color}
\usepackage{amsmath}
\usepackage{xspace}
\newcommand{\zxy}[1]{{\color{blue} [xiangyu: #1]}}

\newcommand{\rc}[1]{{\color{brown} [ruocheng: #1]}}

\usepackage{algorithm}
\usepackage{algorithmic}
\usepackage{subfigure}
\usepackage{enumitem}

\newtheorem{zltheo}{Theorem}
\newtheorem{zlprop}{Proposition}
\begin{document}

\title{Adversarial Curriculum Graph Contrastive Learning with Pair-wise Augmentation}




\author{Xinjian Zhao}
\affiliation{%
 \institution{City University of Hong Kong}
  \streetaddress{1 Th{\o}rv{\"a}ld Circle}
 \city{HongKong}
 \country{China}}
\email{zhaoxinjian2020@gmail.com}

\author{Liang Zhang}
\affiliation{%
 \institution{Shenzhen Research Institute of Big Data}
 \city{Shenzhen}
 \country{China}}
\email{zhangliang@sribd.cn}

\author{Yang Liu}
\affiliation{%
 \institution{Duke Kunshan University}
 \city{Suzhou}
 \country{China}}
\email{yang.liu2@dukekunshan.edu.cn}

\author{Ruocheng Guo}
\affiliation{%
 \institution{ByteDance Research}
 \city{London}
 \country{UK}}
\email{rguo.asu@gmail.com}

\author{Xiangyu Zhao}
\affiliation{%
 \institution{City University of Hong Kong}
  \streetaddress{1 Th{\o}rv{\"a}ld Circle}
 \city{HongKong}
 \country{China}}
\email{xianzhao@cityu.edu.hk}


\begin{abstract}
Graph contrastive learning (GCL) has emerged as a pivotal technique in the domain of graph representation learning. 
A crucial aspect of effective GCL is the caliber of generated positive and negative samples, which is intrinsically dictated by their resemblance to the original data. Nevertheless, precise control over similarity during sample generation presents a formidable challenge, often impeding the effective discovery of representative graph patterns. 
To address this challenge, we propose an innovative framework — Adversarial Curriculum Graph Contrastive Learning (ACGCL), which capitalizes on the merits of pair-wise augmentation to engender graph-level positive and negative samples with controllable similarity, alongside subgraph contrastive learning to discern effective graph patterns therein. 
Within the ACGCL framework, we have devised a novel adversarial curriculum training methodology that facilitates progressive learning by sequentially increasing the difficulty of distinguishing the generated samples. Notably, this approach transcends the prevalent sparsity issue inherent in conventional curriculum learning strategies by adaptively concentrating on more challenging training data. 
Finally, a comprehensive assessment of ACGCL is conducted through extensive experiments on six well-known benchmark datasets, wherein ACGCL conspicuously surpasses a set of state-of-the-art baselines.

\end{abstract}



\maketitle
\input{samples/1Introduction}

\input{samples/3Method}

\input{samples/4Experiments}
\input{samples/2Relatedwork}
\input{samples/5Conclusion}





\clearpage
\bibliographystyle{ACM-Reference-Format}
\bibliography{samples/main}

\clearpage
\appendix

\section{Statistics of datasets}\label{Statistics}
For a comprehensive comparison, we use six widely used datasets, including Cora, Citeseer, Pubmed, Amazon-Computers, DBLP, and Coauthor-Physics, to study the performance of transductive node classification. The datasets are collected
from real-world networks from different domains. The detailed
statistics are summarized in Table \ref{tab:datasets}.
\setcounter{table}{0}
\setcounter{figure}{0}

\begin{table}\centering
	\caption{Statistics of datasets}
	\label{tab:datasets}	\vspace{-3mm}
	\setlength{\tabcolsep}{2.2mm}
	\begin{tabular}{@{} c c c c c @{}}
		\toprule[1pt]
		Dataset  & \#Nodes & \#Edges & \#Features & \#Classes \\
		\midrule
		Cora &  2,708 &  5,429 & 1,433 & 7 \\ 
        Citeseer & 3,327 &4,732& 3,703& 6\\
        Pubmed &  19,717 & 44,338 & 500 &3\\
        Amazon-Computer &13,752 & 245,861 & 767 &10\\
        DBLP& 17,716 &105,734 & 1,639 & 4\\
        Coauthor-Phy& 34,493 & 247,962 & 8,415 &5\\
        \bottomrule[1pt]
	\end{tabular}
		\vspace{-4mm}
\end{table}

\section{Analysis}\label{Analysis}
\subsection{Parameter analysis}\label{ana1} 
\begin{figure}[t]
	\centering
	{\subfigure{\includegraphics[width=0.40\linewidth]{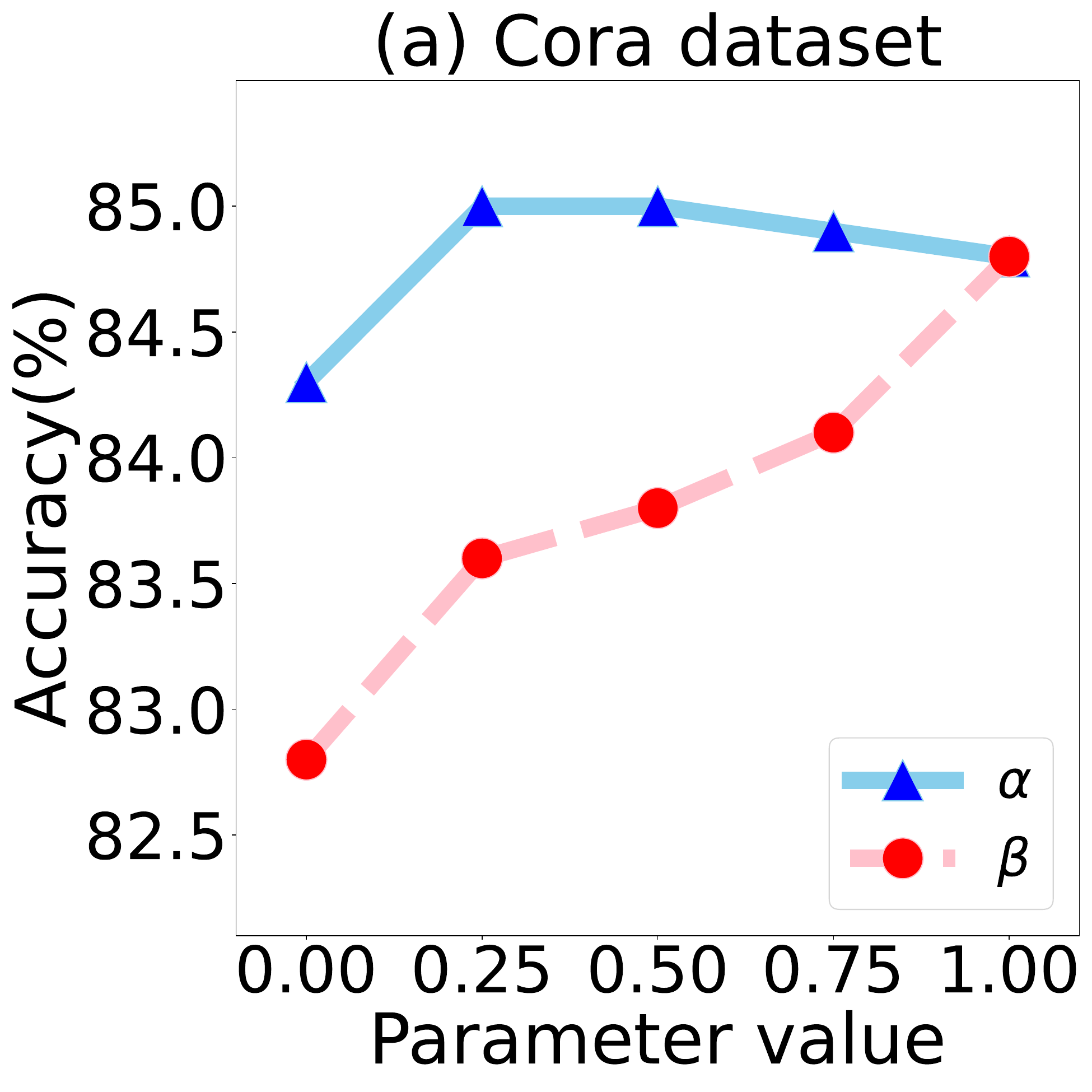}}}
	{\subfigure{\includegraphics[width=0.40\linewidth]{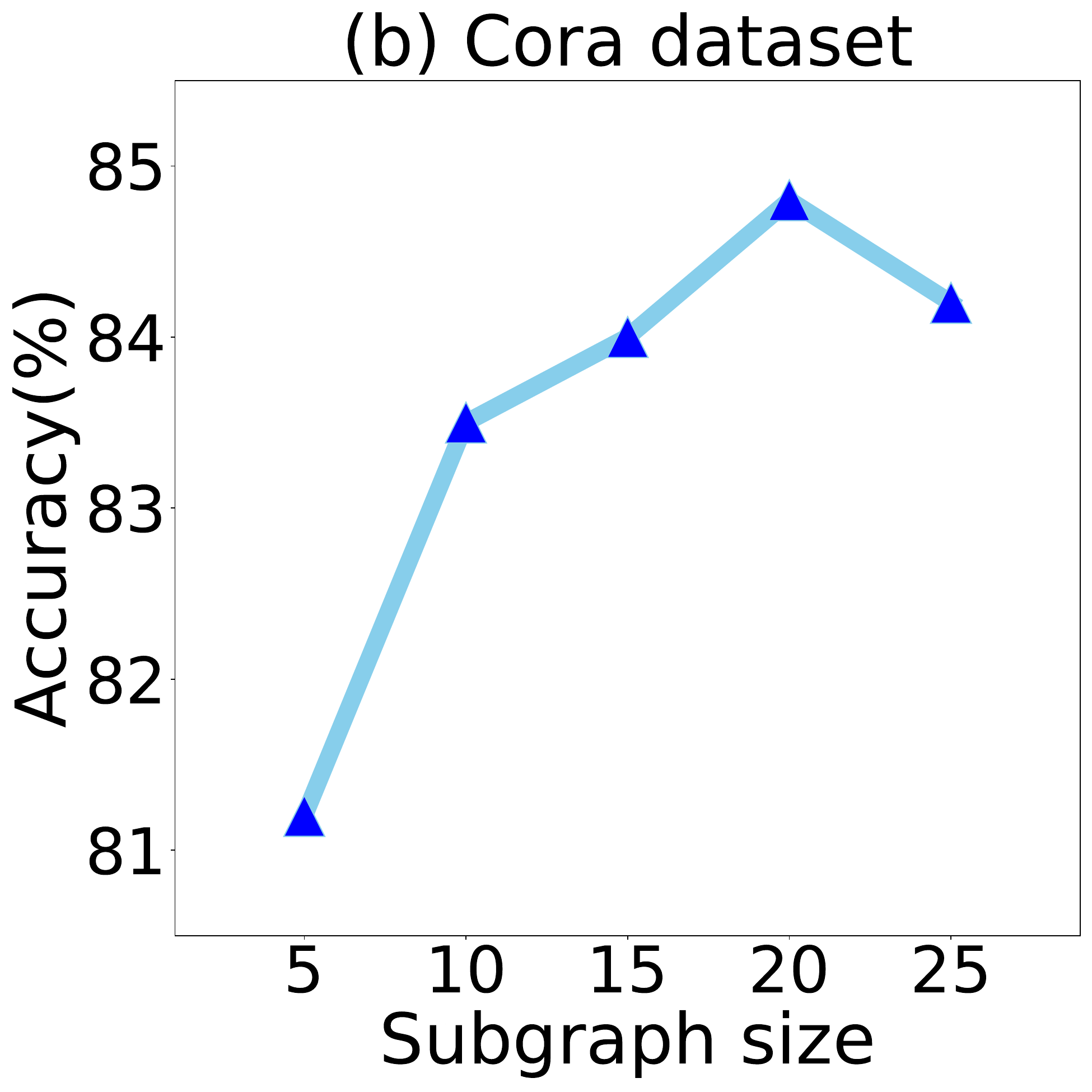}}}	{\subfigure{\includegraphics[width=0.40\linewidth]{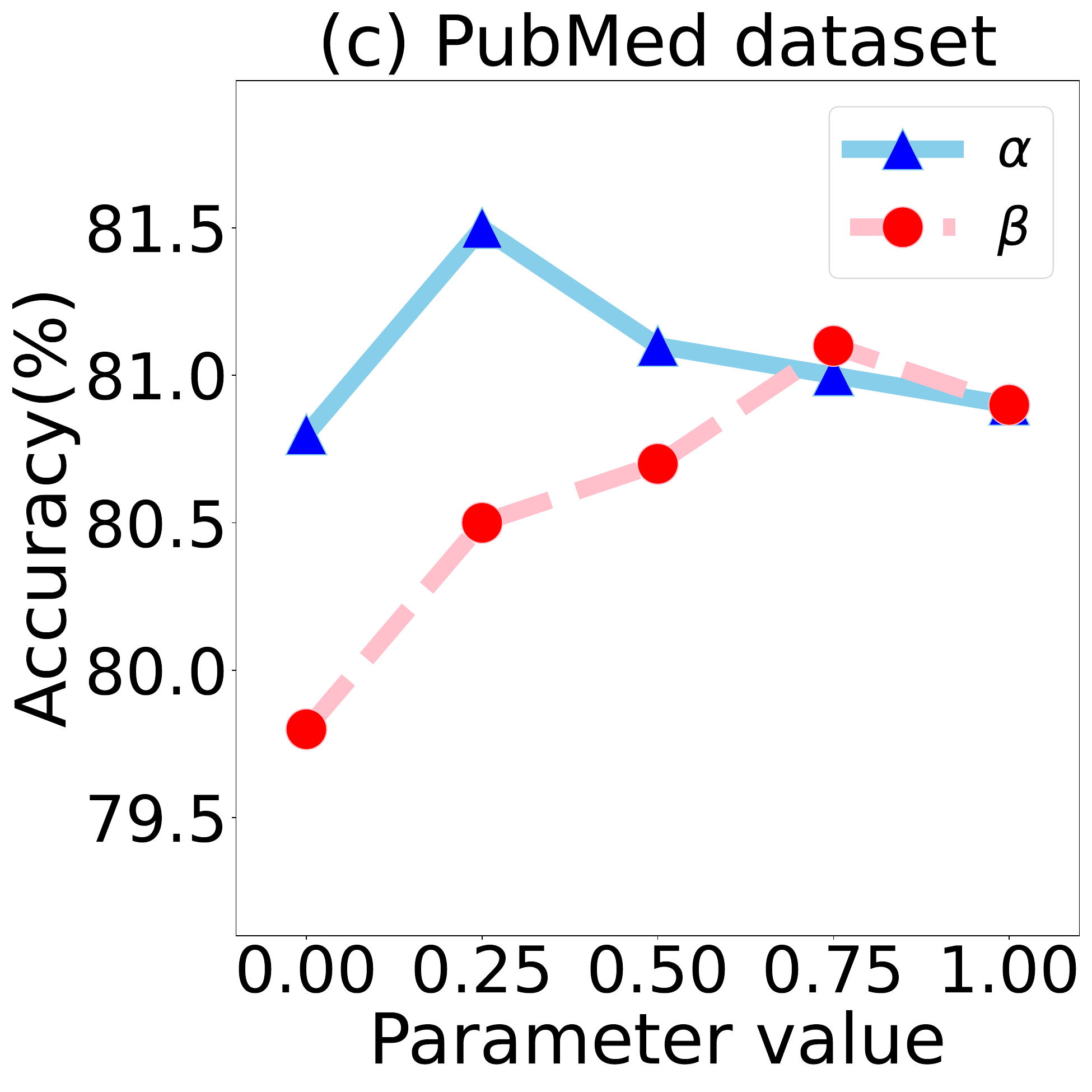}}}	{\subfigure{\includegraphics[width=0.40\linewidth]{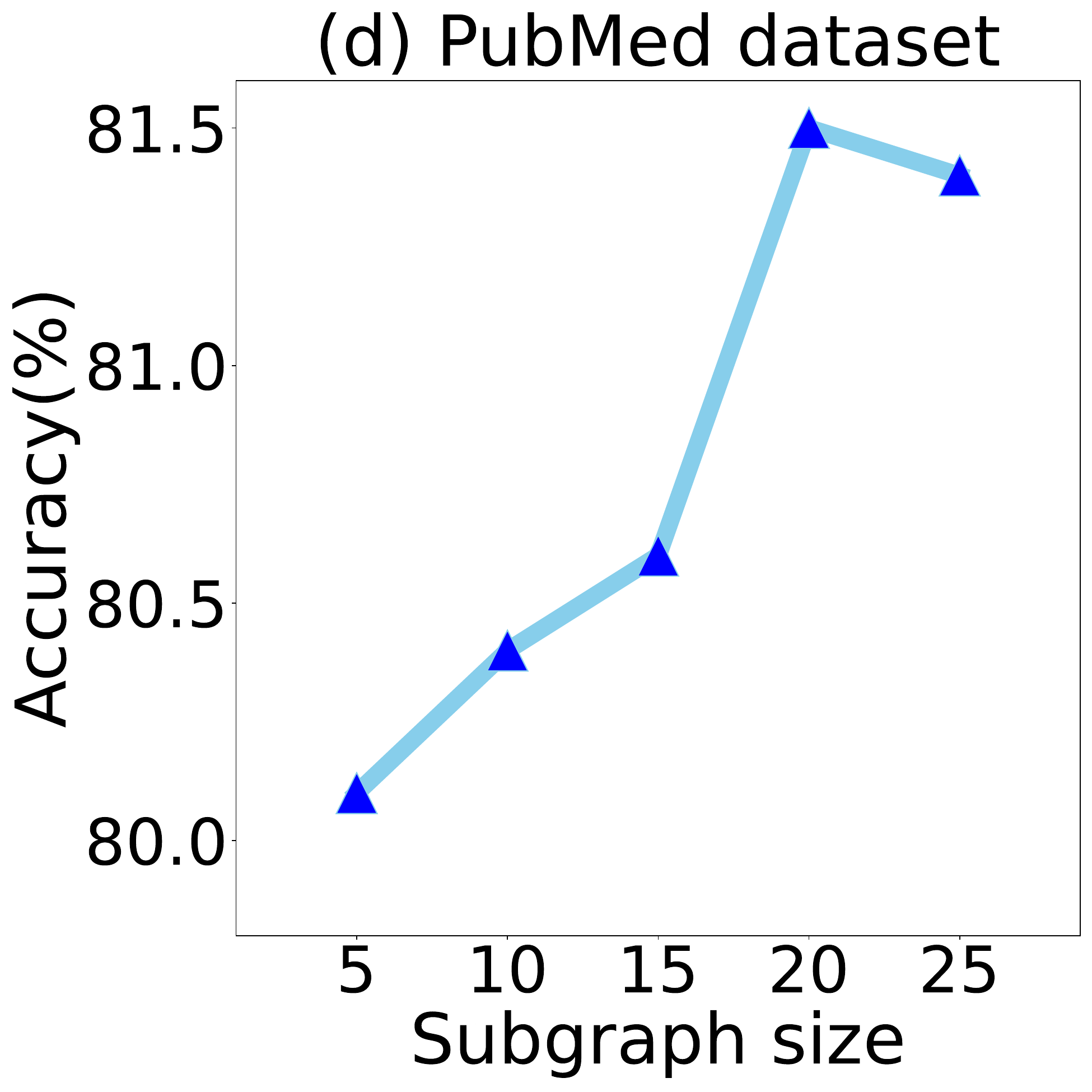}}}
	\vspace{-5mm}
	\caption{Parameter analysis}\label{Parameter}
\end{figure}

Here, we evaluate the robustness of ACGCL under different parameters. We investigate the performance of the model when $\alpha$, $\beta$ in Eq.~(\ref{Loss}), and subgraph size $K$ are varied. The performance of our model with different parameters are shown in Figure \ref{Parameter}.
We find that the weight $\beta$ of $\mathcal{L}_{CL}$ is crucial to the model performance, and choosing an appropriate $\alpha$ helps to improve the performance of the model. When the weight of $\mathcal{L}_{Bal}$ is too large, the node representation distributions of the original subgraphs, positive and negative mirror graphs are too similar to each other, making it difficult to distinguish the original subgraphs from the negative mirror graphs. 
%

\subsection{Distance function selection}\label{ana2}
%
The choice of distance function can have a significant impact on the distribution of node similarity and, in turn, the selection of mirror node pairs. In Table~\ref{distance}, we evaluate the node classification accuracy using different distance functions. We find that the range of values in certain dimensions of the feature vector can be crucial in counterfactual data augmentation. The cosine distance, which only takes into account the direction of the feature vector, results in poor performance. On the other hand, the value-sensitive Euclidean distance and Manhattan distance yield better performance. Additionally, incorporating domain knowledge to weight different dimensions in the node similarity distribution calculation may lead to even better performance, particularly when there is a clear understanding of the dataset.

\begin{table}\centering
	\caption{Effect of different distance functions.}\label{distance}
        \vspace{-3mm}
        \renewcommand
	\label{tab:distance}
	\begin{tabular}{@{} c c c c  @{}}
		\toprule[1pt]
		Dataset  & Euclidean & Cosine & Manhattan \\
		\midrule
		Cora & $\boldsymbol{84.4}$ ± $\boldsymbol{0.6}$  & 83.6 ± 0.4  & 84.1 ± 0.4 \\ 
        Citeseer & $\boldsymbol{73.5}$ ± $\boldsymbol{0.3}$    & 72.6 ± 0.4 & 72.9 ± 0.7  \\
        Pubmed & $\boldsymbol{81.4}$ ± $\boldsymbol{0.2}$ &  80.3 ± 0.5 & 80.4 ± 0.3\\
        \bottomrule[1pt]
	\end{tabular}
\end{table}

\subsection{Subgraph embedding distribution analysis}
\begin{figure}[t] 
\centering 
\includegraphics[width=0.6\linewidth]{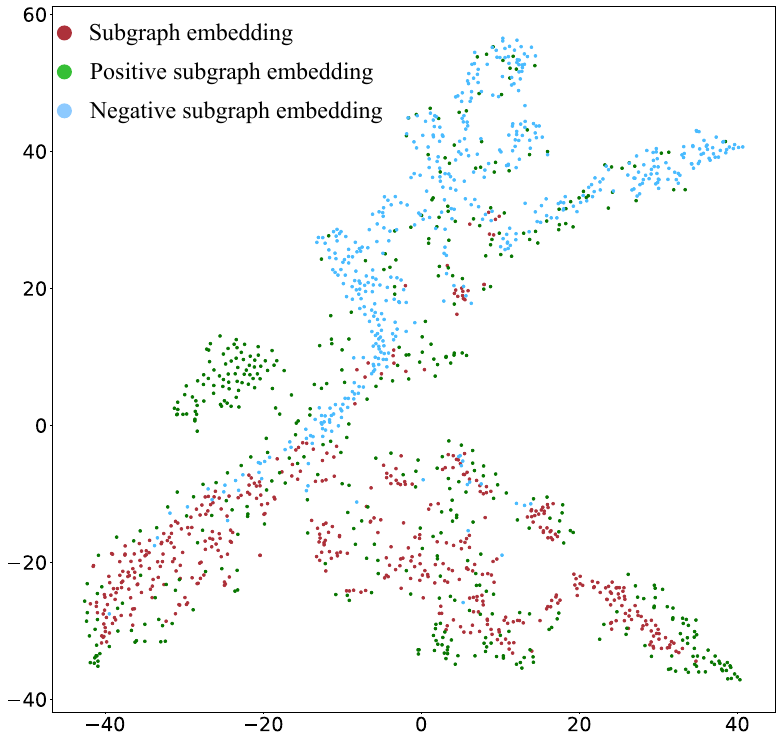} 
\vspace{-5mm}
\caption{Visualization of subgraph embedding distribution}
\label{tsne} 
\end{figure}

Here, we assess whether the distribution of subgraph embeddings aligns with our expectations. In Figure~\ref{tsne}, we employ TSNE~\cite{van2008visualizing} to visualize the embeddings of 500 randomly selected subgraphs and their corresponding positive and negative mirror subgraphs from the Coauthor-Phy dataset. The embeddings are generated using a pre-trained GNN. The results show that in the majority of cases, the embeddings of the original subgraphs are more similar to the positive samples than the negative samples.

\section{algorithm}\label{algorithm}
\begin{algorithm}[t]
\caption{\label{alg:training}  Adversarial Curriculum Learning Strategy}
	\raggedright
        \label{algorithm1}
	{\bf Input}: Hyperparameter $ M , T $; Patience of early stop $p$; Subgraphs $\boldsymbol{S}=\{G_i\}^N_{i=1}$; Pair-wise graph augmentation $PG$; GNN $h_{{\Theta}}$ with learnable parameters $\Theta$; Pacing function $\theta_{\text {linear}}$; Quantile function $q$; Starting epoch $t=0$.\\
	{\bf Output}: GNN $h_{{\Theta}}$ with  learned parameters $\Theta$. \\
	\begin{algorithmic} [1]
		\WHILE {$p>0$}
		\STATE $t \gets t+1$
		\STATE Obtain the difficulty $\theta$ for $PG$: $\theta=\theta_{\text {linear}}(t)$
        \STATE Estimate node pair distance distribution $\mathbf{Dis}$.
        \STATE calculate the $\theta$ quantile values from the distribution of distance $\mathbf{Dis}$: $\gamma=q(\mathbf{Dis},\theta)$.
		\STATE Obtain the augmented mirror subgraphs: $\boldsymbol{S_{+}} , \boldsymbol{S_{-}}=PG(\boldsymbol{S}, \gamma)$
        \STATE Initialize thresholds: $\lambda_{1} , \lambda_{2}$.
		\FOR{step=1 to $M$}
        \STATE Obtain subgraph embeddings:
        
        $\mathbf{H}^{r}, \mathbf{H^{p}},\mathbf{H^{g}}=h_{{\Theta}}\left( \boldsymbol{S}, \boldsymbol{S_{+}} , \boldsymbol{S_{-}}\right)$
		\STATE Calculate the sample loss function $\mathcal{L}^{i}$ by Eq. (\ref{Loss}).
		\STATE Re-weight all samples using closed solution $u_{i}^{*}v_{i}^{*}$ in Eq. (\ref{hardacl}) or Eq. (\ref{softacl}).
		\STATE Obtain the reweighted loss $\mathcal{L}_{Full}$ and optimize using gradient descent, update GNN $h_{{\Theta}}$.
		\STATE$\lambda_{1} \gets \eta_{1} \cdot \lambda_{1}(\eta_{1}>1)$.
        \STATE $\lambda_{2} \gets \eta_{2} \cdot \lambda_{2}(\eta_{2}<1)$.
		\ENDFOR
		\IF{Error on the validation set not decrease}
	     \STATE $p \gets p-1$
		\ENDIF
		\ENDWHILE
	\end{algorithmic}
\end{algorithm}
\noindent In this section, we summarize the algorithmic process in algorithm~\ref{alg:training}.

\section{Theorem}\label{the}
\setcounter{zltheo}{0}
\setcounter{zlprop}{0}
\setcounter{equation}{0}

\begin{zltheo}
If $g(\boldsymbol{v})$ is convex on $[0,1]^N$ and $f(\boldsymbol{u})$ is concave on $[0,1]^N$, then there exists a unique optimal solution for the max-min optimization problem.
\end{zltheo}
\begin{proof}
We separate the optimization of $\boldsymbol{u}$ and $\boldsymbol{v}$ into two sub-optimization problems. We first prove that given $\boldsymbol{v}$, there exists a unique solution for $\boldsymbol{u}$ and vice versa. Then, the optimal solution for $\boldsymbol{v}$ and $\boldsymbol{u}$ can be expressed as a optimal function $\boldsymbol{v} = \mathcal{F}(\boldsymbol{u})$ and $\boldsymbol{u} = \mathcal{G}(\boldsymbol{v})$. Then we prove that these two function has one unique optimal solution, which can get the proof.

We first consider the optimization of $\boldsymbol{v}$ given $\boldsymbol{u}^{*}$ as follows:
\begin{equation}\label{vstar}
\min _{\boldsymbol{v} \in[0,1]^N} \mathcal{L}_1(\boldsymbol{v};\boldsymbol{u}^*)=\sum_i u_{i}^{*} v_{i} \mathcal{L}^i+ \lambda_{1} g\left(\boldsymbol{v}\right)
\end{equation}
The first order derivative of $\mathcal{L}_1(\boldsymbol{v})$ becomes:
\begin{equation}
\frac{\partial \mathcal{L}_1(\boldsymbol{v})}{\partial v_i} = u_{i}^{*}\mathcal{L}^i +  \lambda_{1} \frac{\partial g\left(\boldsymbol{v}\right)}{\partial v_i}
\end{equation}
Since $g(\boldsymbol{v})$ is convex on $[0,1]^N$, then $\frac{\partial g\left(\boldsymbol{v}\right)}{\partial v_i}$ increases with $v_i$. Denote $g_i(v_i)=\frac{\partial g\left(\boldsymbol{v}\right)}{\partial v_i}$. Then, there exist a unique optimal $v_i=g^{-1}_i(-u_{i}^{*}\mathcal{L}^i/\lambda_{1})$ such that $\frac{\partial \mathcal{L}_1(\boldsymbol{v})}{\partial v_i} = 0$. Note that $g^{-1}_i(\cdot)$ is an increasing function, then the optimal $v_i$ decreases with $u_{i}^{*}$.

We then consider the optimization of $\boldsymbol{u}$ given $\boldsymbol{v}^{*}$ as follows:
\begin{equation}\label{ustar}
\max _{\boldsymbol{u} \in[0,1]^N} \mathcal{L}_2(\boldsymbol{u};\boldsymbol{v}^*) = \sum_i u_{i} v_{i}^{*} \mathcal{L}^i+\lambda_{2}f\left(\boldsymbol{u}\right).
\end{equation}
The first order derivative of $\mathcal{L}_2(\boldsymbol{u})$ becomes:
\begin{equation}
\frac{\partial \mathcal{L}_2(\boldsymbol{u})}{\partial u_i} = v_{i}^{*}\mathcal{L}^i +  \lambda_{1} \frac{\partial f\left(\boldsymbol{u}\right)}{\partial u_i}
\end{equation}
Since $f(\boldsymbol{u})$ is concave on $[0,1]^N$, then $\frac{\partial f\left(\boldsymbol{u}\right)}{\partial u_i}$ decreases with $u_i$. Denote $f_i(u_i)=\frac{\partial f\left(\boldsymbol{u}\right)}{\partial u_i}$. Then, there exist a unique optimal $u_i=f^{-1}_i(-v_{i}^{*}\mathcal{L}^i/\lambda_{2})$ such that $\frac{\partial \mathcal{L}_2(\boldsymbol{u})}{\partial u_i} = 0$. Note that $f^{-1}_i(\cdot)$ is an decreasing function, then the optimal $u_i$ increases with $v_{i}^{*}$.

Note that optimal $\boldsymbol{u}$ and $\boldsymbol{v}$ should be bounded in range $[0,1]^N$. If the optimal solution is not on the bound, then there exists a unique optimal solution since optimal $u_i$ increases with $v_{i}^{*}$ and optimal $v_i$ decreases with $u_{i}^{*}$. Then, we consider there exists an optimal solution on the bound. Here we only analyze the optimal $u_i$ and the $v_i$ can be analyzed in the same way. If the optimal $v_{i}$ is equal to 0, then $u_{i}$ should be on 0, 1 or $f^{-1}_i(0)$. If $f^{-1}_i(0)$ is in range [0,1], then $f^{-1}_i(0)$ is the optimal solution. otherwise, 0 or 1 should be the optimal. Note that 0 or 1 cannot be the optimal solution at the same time. A similar proof can be obtained when the optimal $v_{i}$ is equal to 1. Then, we get the proof.

\end{proof}
\begin{zlprop}
If $g(\boldsymbol{v})$ is convex on $[0,1]^N$ and $f(\boldsymbol{u})$ is concave on $[0,1]^N$, then the optimal solution $v_i^*$ decreases with $\mathcal{L}^i$ and $u_i^*$ increases with $v_i^*\mathcal{L}^i$. In addition, the optimal solution $u_i^*$ decreases with $\lambda_2$ and $v_i^*$ increases with $\lambda_1$.
\end{zlprop}
\begin{proof}
In the proof of Theorem 1, we have proven that the optimal solution should satisfy $u^*_i=f^{-1}_i(-v_{i}^{*}\mathcal{L}^i/\lambda_{2})$ and $v^*_i=g^{-1}_i(-u_{i}^{*}\mathcal{L}^i/\lambda_{1})$. Note that $g^{-1}_i(\cdot)$ is an increasing function and $f^{-1}_i(\cdot)$ is a decreasing function. Obviously, the optimal solution $u_i^*$ decreases with $\lambda_2$ and increases with $v_i^*\mathcal{L}^i$, and $v_i^*$ increases with $\lambda_1$. Then, we consider the case of increasing $\mathcal{L}^i$. If we assume that $v^*_i$ is also increased, then $u^*_i$ should be also increased according to the first optimal equation. According to the second optimal equation, $v^*_i$ should be decreased. This is not possible according to the assumption. Then, $v^*_i$ should decrease with $\mathcal{L}^i$.
\end{proof}
\begin{zlprop}
If the regular functions are defined in Eq.~[\ref{eq:hard_label}]. Then, the optimal solution $u_{i}^{*}, v_{i}^{*}$ satisfy:
\begin{equation}
u_{i}^{*}v_{i}^{*}=\left\{\begin{array}{lr}
1, &  \lambda_{2}<\mathcal{L}^{i}<\lambda_{1}; \\
0, & \text {otherwise}.
\end{array}\right.
\label{hardacl}
\end{equation}
\end{zlprop}
\begin{proof}
If $g(\boldsymbol{v})$ is convex on $[0,1]^N$ and $f(\boldsymbol{u})$ is concave on $[0,1]^N$, then there exists a unique optimal solution for the above max-min optimization problem.
Given: 
\begin{equation}
g(\boldsymbol{v} ; \lambda_{1})=-\lambda_{1} \sum_{i=1}^{N} v_{i}, \quad f(\boldsymbol{u} ; \lambda_{2})=-\lambda_{2} \sum_{i=1}^{N} u_{i}.
\end{equation}
Fix $u$: 
\begin{equation}\label{vstar}
v_{i}^{*}=\arg \min _{v_{i} \in[0,1]} u_{i}^{*} v_{i} \mathcal{L}^i - \lambda_{1} v_{i}, \quad i=1,2, \cdots, N;
\end{equation}
the partial derivative of $v_{i}^{*}$ is:
\begin{equation}
    \frac{\partial \left(u_i^* v_i \mathcal{L}^i - \lambda_1 v_i\right)}{\partial v_i} =u_i^* \mathcal{L}^i -\lambda_1
\end{equation}
We learn the global optimum $v_{i}^{*}$ that if $u_i^* =0$, $v_i^* =1$, else if $u_i^* \neq 0$: 

\begin{equation}\label{vi}
v_{i}^{*}=\left\{\begin{array}{lr}
1, &  \mathcal{L}^{i}<\frac{\lambda_{1}}{u_i ^*}; \\
0, & \text {otherwise}.
\end{array}\right.
\end{equation}
Similarly, we can get the global optimum $u_{i}^{*}$ that if $v_i^* =0$, $u_i^* =0$, else if $u_i^* \neq 0$: 
\begin{equation}\label{u_i}
u_{i}^{*}=\left\{\begin{array}{lr}
0, &  \mathcal{L}^{i}<\frac{\lambda_{1}}{v_i ^*}; \\
1, & \text {otherwise}.
\end{array}\right.
\end{equation}
Now, consider the combined conditions of Eq.~\ref{vi} and Eq.~\ref{u_i}, where the values of $v_i^*$ and $u_i^*$ are interrelated. If $v_i^* = 0$, then according to Eq.~\ref{u_i}, $u_i^*$ must be 0. Similarly, if $u_i^* \neq 0$, then according to Eq.~\ref{vi}, $v_i^*$ must be 1. Therefore, the values of $u_i^*$ and $v_i^*$ are mutually determined.
:
\begin{equation}\label{uv_i}
u_{i}^{*}v_{i}^{*}=\left\{\begin{array}{lr}
1, &  \lambda_{2}<\mathcal{L}^{i}<\lambda_{1}; \\
0, & \text {otherwise}.
\end{array}\right.
\end{equation}
This implies that $u_i^*v_i^*$ equals 1 only when both $u_i^* = 1$ and $v_i^* = 1$. From the derivation of Eq.~\ref{vi} and Eq.~\ref{u_i}, we know that $u_i^*$ and $v_i^*$ are interrelated binary variables, so $u_i^*v_i^*$ equals 1 only when both of them are 1. This completes the derivation from Eq.~\ref{vi} and Eq.~\ref{u_i} to Eq.~\ref{uv_i}. This gives us the hard version of the ACL.
\end{proof}

\end{document}

%% file: samples/1Introduction.tex
\section{Introduction}

Graph is among the commonest non-Euclidean data structures that have been used in a wide range of Web-based applications such as social networks~\cite{hamilton2017inductive}, citation networks~\cite{tang2008arnetminer}, and recommender systems~\cite{wu2019session,wu2022graph,gao2023survey,zhao2023embedding}. To accurately capture the characteristics of the graph structure, graph representation learning algorithms have been proposed to learn low-dimensional representations of nodes or graphs by hierarchically aggregating information in their neighborhoods. The effectiveness of these algorithms have been demonstrated across a variety of downstream tasks including node classification~\cite{kipf2016semi}, link prediction~\cite{zhang2018link}, and graph classification~\cite{lee2019self}, which are further applied in related practical scenarios such as drug discovery~\cite{shi2020graph} and personalized recommendation~\cite{wei2019mmgcn,guan2022bi}. Among the existing graph representation learning techniques, graph contrastive learning (GCL) has gained significant popularity due to its advantages in extracting informative graph representations.

As a core component of GCL, contrastive learning~\cite{chen2020simple,khosla2020supervised} has emerged as a powerful technique for representation learning by minimizing the distance between positive samples and maximizing it between positive and negative samples in the representation space. However, in GCL, the complex latent semantic (e.g., graph class label, topology structure) of graphs poses a significant challenge for generating or sampling high-quality positive and negative samples~\cite{lee2022augmentation}. 
%
Methods like GraphCL \cite{you2020graph} and CuCo \cite{chu2021cuco} perform GCL on the graph-graph level.
They construct positive samples of a graph with rule-based graph augmentation methods such as node dropping and edge perturbation etc.
%
Contrastive learning at the node-graph level has been explored in previous work \cite{hjelm2018learning, peng2020graph, jiao2020sub}.
DGI \cite{hjelm2018learning} constructs positive samples by aggregating node representations of the original graph and negative samples by shuffling node ordinal numbers. GMI \cite{peng2020graph} and SUBG-CON \cite{jiao2020sub} propose more efficient GCL methods, they construct positive and negative samples from local subgraphs of each node. Therefore, it is crucial to develop advanced GCL methods to address this challenge and achieve efficient representation learning for graph-based data.

Semantic similarity and distance in representation space are the two key factors of positive and negative samples that determine the effectiveness of a GCL method~\cite{xia2022progcl,liu2022revisiting}.
Positive samples are desired to be semantically similar. Simultaneously, it is crucial to ensure that a negative sample has significantly different latent semantics from a positive one but maintains a proper distance in the representation space~\cite{xia2022progcl,ying2018graph,yang2020understanding}.
%
%
However, existing GCL methods can hardly control the semantic similarity and the distance in the representation space between the original samples and the corresponding positive and negative samples. This is because they generate positive and negative samples from a limited space of graphs, such as sampled subgraphs, or through predefined perturbations.
Thus, generated positive samples might not be similar enough to the original sample in terms of semantic meaning.
For example, in DGI~\cite{velickovic2019deep}, a positive sample is defined as the mean representation of all nodes in the original graph. Such positive samples might not be proper for those nodes with unique characteristics, for example, users from a minority community in social networks.
In addition, most of the existing methods obtain negative samples by random sampling from the original graph,
without measuring the semantic similarity between positive and negative samples. Thus, in different stages of training, the negative samples can either be too easy to distinguish from the positive ones, or too hard as they are semantically similar to positive samples and prone to be false negatives.

In order to tackle the aforementioned challenges, an innovative \textbf{A}dversarial \textbf{C}urriculum \textbf{G}raph \textbf{C}ontrastive \textbf{L}earning (ACGCL) framework is proposed in this work, in which a set of strategies are developed to accommodate the corresponding issues. In specific, a novel \textbf{pair-wise graph augmentation} method is developed to generate positive and negative samples with both desired semantic characteristics and controllable distances in the representation space.
Unlike conventional graph augmentation methods (e.g., node dropping~\cite{yin2022autogcl} and graph generation~\cite{ma2022learning,zhao2021counterfactual}) that can only create positive samples from the graph level, our method leverages the advantages of GCL to simultaneously craft a pair of positive and negative graph structures termed as mirror graphs.
Specifically, given a pair of nodes in a graph, our method finds another pair of nodes that is sufficiently similar (i.e., distance in representation space lower than pre-defined threshold) to the original pair as the mirror pair.
Then these mirror pairs are used to generate neighbor relationships for new mirror graphs. When the mirror pairs share identical local semantics, such as the article categories of nodes in citation networks, a positive mirror graph is created by substituting the neighbor relationships of the original pairs with those of the mirror pairs. Conversely, if the two pairs possess different local semantics, a similar approach is applied to adjust the neighbor relationships of the original graph, resulting in the creation of a negative mirror graph. Consequently, the positive samples display a strong resemblance, whereas the negative samples exhibit a weaker similarity to the original graph.
To reduce the complexity of learning graph representations in the large graph, we pre-sample a large original graph into a set of sub-graphs with appropriate sizes, inspired by previous works~\cite{hassani2020contrastive,qiu2020gcc,jiao2020sub,zhang2020motif}, and apply pair-wise graph augmentation on each of them, 
Furthermore, a \textbf{subgraph contrastive learning} method is developed, which investigates the specific characteristics of the mirror graphs to learn informative inter-graph and intra-graph patterns.

To make full use of the various difficulty levels defined by the semantic similarities of generated graph samples, curriculum learning~\cite{bengio2009curriculum} is employed to construct a progressive training framework, which enables effective model training through mimicking the way human learns, i.e., gradually increasing the difficulty of training data by excluding difficult samples at the early training stages and adding them back as the training proceeds.
While existing curriculum learning methods \cite{hacohen2019power,kumar2010self} are limited by the sparsity of difficult samples, a novel \textbf{adversarial curriculum training} framework is proposed in this work, which utilizes a predefined pacing function to control the difficulty of the samples generated through pair-wise graph augmentation. To achieve the accurate control of difficulty, it only utilizes samples with losses within a reasonable range for training in each epoch. Furthermore,  to alleviate the sparsity problem of difficult samples, larger weights can be assigned to difficult samples via
adversarial learning, enabling the model to gradually focus on the difficult samples.

In summary, our main contributions can be demonstrated as:
\begin{itemize}[leftmargin=*]
\item We proposed a novel pair-wise graph augmentation method for generating positive and negative samples with controllable similarity and a sub-graph contrastive learning method to extract effective graph patterns from augmented graphs.  \
\item We propose a novel adversarial curriculum training framework to address the sparsity issue in curriculum learning. \
\item We conduct extensive experiments to verify the effectiveness of the proposed framework on multiple widely used benchmark datasets for node classification.
\end{itemize}

%% file: samples/3Method.tex
\section{METHODOLOGY}
\begin{figure*}[t] 
\centering 
\includegraphics[width=0.95\textwidth]{samples/images/pipeline-3.pdf} 
\vspace{-3mm}
\caption{Overview of the proposed framework ACGCL} 
\label{Fig.Over} 
\vspace{-3mm}
\end{figure*}
In this section, we propose a novel framework ACGCL for learning effective node representations for downstream tasks such as node classification. ACGCL consists of three main components: 1) pair graph augmentation; 2) subgraph contrastive learning; 3) adversarial curriculum training. In pair-wise graph augmentation, we find sufficiently similar node pairs in the graph as mirror node pairs, and subsequently, we replace the links between the original nodes with the help of mirror node pairs to obtain new positive and negative mirror graphs. To avoid the high complexity caused by inputting the complete graph for more efficient training, we apply pair-wise graph augmentation to subgraph contrastive learning. Finally, we design an adversarial curriculum training framework that couples well with pair-wise graph augmentation, allowing the model to learn from easy to hard and giving more reasonable weights to difficult samples. An overview of the ACGCL is shown in Figure \ref{Fig.Over}.

\subsection{Preliminaries}

We let $\mathcal{G}=(\mathcal{V},\mathcal{E})$ denote a graph, where $\mathcal{V}=\left\{v_{1}, v_{2}, \cdots, v_{N}\right\}$, $
\mathcal{E} \subseteq \mathcal{V} \times \mathcal{V}
$ are the set of nodes and edges of the graph, respectively. Each node in the graph has $D$-dimensional features, and the feature matrix can be denoted as $\mathbf{X}=\left\{\mathbf{x}_{1}, \mathbf{x}_{2}, \ldots, \mathbf{x}_{N}\right\}\in \mathbb{R}^{N \times D}$.   The adjacency matrix of a graph is: $\mathbf{A} \in\{0,1\}^{N \times N}$. In this work, we focus on the undirected graph. If there is a link between $v_{i}$ and $v_{j}$ , $a_{ij}= a_{ji} = 1$, otherwise $a_{ij}=a_{ji}= 0$. In this work, we aim to learn a GNN encoder $f:\mathbb{R}^{N \times D}\times\{0,1\}^{N \times N} \rightarrow \mathbb{R}^{N \times d}, d\ll D$ that maps the feature matrix $\mathbf{X}$ and adjacency matrix $\mathbf{A}$ to low-dimensional node representations $\mathbf{H} \in \mathbb{R}^{N \times d}$ s.t. $\mathbf{H}$ provides highly discriminative features for node classification.

\noindent \textbf{Subgraph Sampling.} Sampling subgraphs for graph learning is a widely adopted method to reduce the high complexity. The studies \cite{jiao2020sub,hamilton2017inductive} point out that nodes are mainly influenced by a small number of nearby nodes. Compared with the methods directly utilizing n-hop subgraphs, the personalized pagerank (PPR) \cite{klicpera2018predict} algorithm has been verified to be more efficient.The importance score matrix $\mathbf{S}$ is computed by:
\begin{equation}\label{equation-1}
\mathbf{S}=\alpha \cdot(\mathbf{I}-(1-\alpha) \cdot \overline{\mathbf{A}}),
\end{equation}
where $\overline{\mathbf{A}}=\mathbf{A} \mathbf{D}^{-1}$denotes the column-normalized adjacency matrix.

For each node $i$, we find the $K$ nodes with the highest score from the score matrix $\mathbf{S}$, and establish its feature matrix $\mathbf{X_{i}}$ and adjacency matrix $\mathbf{A_{i}}$. This method can greatly reduce memory consumption and aggregate neighborhood information more effectively. More formally, the process can be written as:
\begin{equation}\label{equation-2}
\mathcal{I}_i=toprank\big(\mathbf{S}(i,:), K\big),
\end{equation}
\begin{equation}\label{equation-3}
\mathbf{X}_{i}=subgraph(\mathbf{X}, \mathcal{I}_i), \quad \mathbf{A}_{i}=subgraph(\mathbf{A},\mathcal{I}_i).
\end{equation}

\noindent \textbf{Self-paced Learning.} The self-paced learning strategy (SPL) \cite{kumar2010self} utilizes the loss value to measure the sample difficulty, which can dynamically evaluate the difficulty of different samples at different steps. The objective function of SPL can be formulated as:
\begin{equation}
\min _{\boldsymbol{w} ; \boldsymbol{v} \in[0,1]} \mathbb{E}(\boldsymbol{w}, \boldsymbol{v} ; \lambda)= \sum_{i=1}^{N} v_{i} l_{i}(\boldsymbol{w})+g(\boldsymbol{v} ; \lambda),
\end{equation}
where $\boldsymbol{w}$ denotes the model parameters. $v_{i}$, $l_{i}$ are the weight and the loss of sample $i$. $g(\boldsymbol{v} ; \lambda)= -||\boldsymbol{v}||_1$ is a negative $L_1$-norm of sample weights. SPL can obtain a closed optimal solution $\boldsymbol{v^{*}}$ with fixed $\boldsymbol{w}$:
\begin{equation}
v_{i}^{*}=\left\{\begin{array}{lr}
1, & l_{i}<\lambda; \\
0, & \text { otherwise. }
\end{array}\right.
\end{equation}
With $\boldsymbol{v}$ fixed, $\boldsymbol{w}$ can be optimized by gradient descent.
%
\subsection{Pair-wise Graph Augmentation} 
Previous graph contrastive learning methods~\cite{hassani2020contrastive,you2020graph,jiao2020sub} mostly focus on how to define positive samples and only use a naive random method to sample negative samples. These methods cannot generate high-quality positive and negative samples with controllable difficulty.
Here, we describe the pair-wise graph augmentation framework to obtain positive and negative samples with controllable similarities. Our proposed pair-wise graph augmentation modifies the adjacency matrix of the original graph with mirror node pairs.
Specifically, for a pair of nodes $v_{a},v_{b}$ in the original graph, another pair of nodes with high semantic similarity (dissimilarity) to them can be their mirror pairs. 
Our pair-wise graph augmentation is partially inspired by the work \cite{zhao2021counterfactual}. Different from this work which defines selected node pairs only based on feature similarity, we propose to ensure high semantic similarity (dissimilarity) between the original pair and its mirror pairs by imposing extra constraints w.r.t. a semantic variable. The intuition about the semantic variable is that if we change the semantic variable, the graph structure can be changed significantly. The semantic variable can be the degree of each node or the class label. Other semantic variables like the eigenvalue from the graph spectral view can also be considered. Formally, we define the semantic variables as $\mathcal{Y}=\left\{y_{1}, y_{2}, \cdots, y_{N}\right\}$.
%
%
%


\begin{figure}[t] 
\centering 
\includegraphics[width=0.35\textwidth]{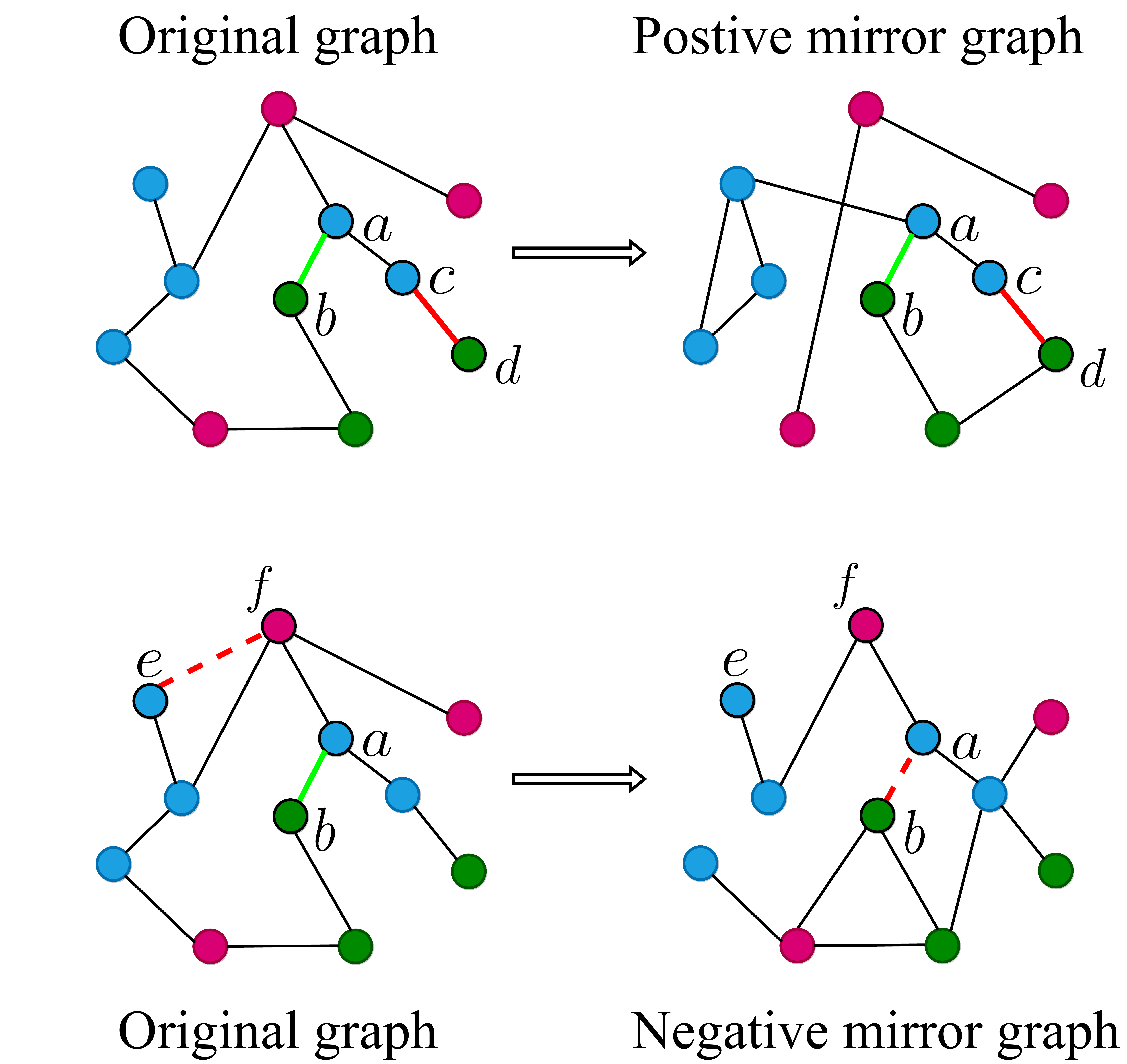}
\vspace{-3mm}
\caption{An example of one step in pair-wise graph augmentation. Let a solid (dashed) edge denote that there is a (no) edge between the two nodes. Given a pair of nodes $v_a$ and $v_b$ in the original graph, their positive mirror pair $v_c$ and $v_d$ and negative mirror pair $v_e$ and $v_f$, the green edge between node $v_{a}$ and $v_{b}$ in the original graph is maintained to create the positive mirror graph as the positive mirror pair $v_c$ and $v_d$ are connected. While it will be removed in the negative mirror graph as there is no edge between the pair $v_e$ and $v_f$.} 
\label{Fig.CF} 
\vspace{-4.9mm}
\end{figure}


 Given a graph $\mathcal{G}$, we refer to the positive mirror graph as $\mathcal{G}^+=(\mathcal{V},\mathcal{E}^+)$ and negative mirror graph as $\mathcal{G}^-=(\mathcal{V},\mathcal{E}^-)$, where $\mathcal{E}^+$ and $\mathcal{E}^-$ are the edge set of them.
 Note that the mirror graphs share the same set of nodes and node features as the original graph $\mathcal{G}$, but different edge sets.
 For a given node pair $(v_{a},v_{b})$ on the original graph $\mathcal{G}$, its positive and negative mirror pairs are signified as $(v^+_{a},v^+_{b})$ and $(v^-_{a},v^-_{b})$. The mirror pairs are selected from the original graph $\mathcal{G}$. Nodes in a mirror pair are the nearest neighbors of the original pair in the feature space with extra constraints. In particular, nodes in a positive (negative) mirror pair have the same label (different labels) as (from) the original pair. Formally, given a pair of nodes $(v_{a},v_{b})$ in the original graph, their positive and negative mirror pairs are obtained as follows:
\begin{equation} \label{equation-4}
\begin{split}
(v^+_{a}, v^+_{b})=\underset{v_{c}, v_{d} \in \mathcal{G}/\{v_{a}, v_{b}\}}{\arg \min }\big\{d({\mathbf{x}}_{c}, {\mathbf{x}}_{a})+d({\mathbf{x}}_{d}, {\mathbf{x}}_{b})
\mid y^{a}=y^{c}\\
\cap \;y^{b}=y^{d},d({\mathbf{x}}_{c}, \mathbf{x}_{a})+d({\mathbf{x}}_{d}, {\mathbf{x}}_{b})<2 \gamma \big\},
\end{split}
\end{equation}

\begin{equation}\label{equation-5}
\begin{split}
(v^-_{a}, v^-_{b})=\underset{v_{e}, v_{f} \in \mathcal{G}/\{v_{a}, v_{b}\}}{\arg \min }\big\{d({\mathbf{x}}_{e}, {\mathbf{x}}_{a})+d({\mathbf{x}}_{f}, {\mathbf{x}}_{b}) 
\mid y^{a} \neq y^{e}\\
\cup \; y^{b}\neq y^{f},d({\mathbf{x}}_{e}, {\mathbf{x}}_{a})+d({\mathbf{x}}_{f}, {\mathbf{x}}_{b})<2 \gamma\big\},
\end{split}
\end{equation}
where $d(\cdot,\cdot)$ denotes the distance function (e.g., the Euclidean distance) and $2\gamma$ is the threshold on the distance between node pairs determining whether a mirror node pair would be considered.
An example of mirror graph augmentation for a given node pair $v_ {a},v_ {b}$ is illustrated in Figure \ref{Fig.CF}.

Since we use the same semantic mirror pairs as the original node pairs to generate adjacency matrices for the positive mirror graphs, the structural semantics of the positive mirror graphs are closer to the original graphs, and vice versa for the negative mirror graphs. Note that we can control the similarity between mirror node pairs and the original ones through the hyperparameter $\gamma$ in Eq.~\eqref{equation-4}. We denote the proposed pair-wise graph augmentation strategy as:
\begin{equation}
 (\mathbf{A_{+}},\mathbf{A_{-}})=CG(\mathbf{A},\mathbf{X},\gamma),
\end{equation}
where $CG$ is the pair-wise graph augmentation operator, and $\mathbf{A_{+}}$ ($\mathbf{A_{-}}$) are the adjacency matrices of the positive (negative) mirror graph of the original graph $\mathbf{A}$. After mirror graph augmentation, the original graph $\mathcal{G}$, the positive mirror graph $\mathcal{G}^+$ and the negative mirror graph $\mathcal{G}^-$ can be used for contrastive learning. The positive mirror graph is a good positive sample in GCL since it has high semantic and structural similarity to the original graph.
While the semantics and structure of negative samples are corrupted, we can change the degree of corruption by adjusting parameter $\gamma$ in Eq. (\ref{equation-5}).
We adopt a GNN encoder $h_{{\Theta}}$ to update the representation of a node by aggregating the information
of its neighbors. The updated representation of node $i$ at layer $l+1$ can be formulated as follows:


\begin{equation}
h_i^{(l+1)}=\operatorname{UPDATE}\left(h_i^{(l)}, \operatorname{AGG}\left(\left\{h_u^{(l)} \mid u \in \mathcal{N}(i)\right\}\right)\right)
\end{equation}
where $h_v^{(l)}$ denotes the representation of node $i$ at layer $l$, $\operatorname{AGG}$ and $\operatorname{UPDATE}$ denote the aggregate and update operator, respectively, and $\mathcal{N}(i)$ denotes the set of neighbors of node $i$. $\mathbf{H}^{(l)}$ denotes the node representation matrix as the output of the $l$-th layer. 
Through the GNN encoder $h_{{\Theta}}$, we can get the representation matrices of the original graph and its positive (negative) graph :
\begin{equation}\label{eq:total_emb}
\mathbf{H}=h_{{\Theta}}(\mathbf{A},\mathbf{X}), ~~ \mathbf{H^{+}}=h_{{\Theta}}(\mathbf{A_{+}},\mathbf{X}), ~~ \mathbf{H^{-}}=h_{{\Theta}}(\mathbf{A_{-}},\mathbf{X}).
\end{equation}

In this paper, we learn the graph representations via a contrastive learning method. We intend to minimize the distance between the original graph $\mathcal{G}$ and positive mirror graph $\mathcal{G}^+$, and maximize the distance between the original graph's embedding $\mathcal{G}$ and negative mirror graph $\mathcal{G}^-$. Note that we measure the graph distance from the node level. We adopt the margin triple loss \cite{schroff2015facenet} function as the optimization objective of contrastive learning:
\begin{equation}
\mathcal{L}_{CL}=\sum_{i=1}^{N}\left[\left\|\mathbf{h_{i}}-\mathbf{h_{i}^{+}}\right\|_{2}^{2}-\left\|\mathbf{h_{i}}-\mathbf{h_{i}^{-}}\right\|_{2}^{2}+\xi\right]_{+},
\end{equation}
where $\xi$ is margin value, $\mathbf{h}_{i},{\mathbf{h}_{i}^{+}}$ and ${\mathbf{h}_{i}^{-}}$ represent the embedding of node $i$ in the three graphs from Eq.~\eqref{eq:total_emb}.

In the process of pair-wise graph augmentation, we need to maintain a distance matrix between nodes. The space complexity of pair-wise data augmentation is $O(N^{2})$ and grows rapidly as the number of nodes grows. Therefore, pair-wise graph augmentation is not scalable to large graphs. Fortunately, it has been shown that contrastive learning on local subgraphs can learn highly effective node representations. Thus, we sample the large graph into a set of subgraphs and implement pair-wise graph augmentation and contrastive learning for all subgraphs.


\subsection{Subgraph Contrastive Learning}

As mentioned above, pair-wise graph augmentation can suffer from the high time and space complexity of large graphs. In order to effectively cope with our contrastive learning framework with large graphs, the subgraph sampling method mentioned in Section 2.1 is adopted. More specifically, we sample the original graph $\mathcal{G}$ in advance into a set of subgraphs denoted as $\{G_i\}^N_{i=1}$, where $G_i$ is the subgraph for node $i$. Based on the original subgraphs, we generate $N$ positive mirror subgraphs $\{G^+_i\}^N_{i=1}$ and $N$ negative mirror subgraphs $\{G^-_i\}^N_{i=1}$ accordingly. Note that for each node $i$, the subgraph $G_i$, the positive subgraph $G^+_i$ and the negative subgraph $G^-_i$ have the same node set $\mathcal{I}_i$ according to the subgraph sampling in Eq.~\eqref{equation-2}. In addition, we let all the subgraphs share the same size, i.e., $|\mathcal{I}_i|=K$, leading to constant time and space complexity $O(K^2)=O(1)$. Thus, the pair-wise data augmentation carried out on the subgraphs enjoys significant improvement in the efficiency of the algorithm w.r.t. time and space.

\noindent \textbf{Inter-graph contrastive loss.} As introduced in Section 2.2, the contrastive learning loss among the original graph and the augmented graphs can be found in Eq.~\eqref{eq:total_emb}. Here, we extend it for contrastive learning loss on subgraphs. For subgraph $G_i$, its inter-graph contrastive learning loss is defined as:
\begin{equation}
\mathcal{L}_{C L}^i=\frac{1}{K} \sum_{j \in I_i}\left[\left\|\mathbf{h}_{\mathbf{j}}-\mathbf{h}_{\mathbf{j}}^{+}\right\|_2^2-\left\|\mathbf{h}_{\mathbf{j}}-\mathbf{h}_{\mathbf{j}}^{-}\right\|_2^2+\xi\right]_{+} .
\end{equation}
Given the inter-graph contrastive learning loss $\mathcal{L}^i_{CL}$ for each subgraph $G_i$, the final contrastive learning loss can be written as:
\begin{equation}
\mathcal{L}_{CL}= \frac{1}{N} \sum_{i=1}^{N} \mathcal{L}_{CL}^{i}.
\end{equation}

\noindent \textbf{Batch balance loss.} 
In contrastive learning, ensuring similar representation distributions for positive and negative samples is vital to prevent the model from easily distinguishing between them, thus avoiding model collapse. Additionally, maintaining the similarity between the distribution of node representations in mirrored graphs and the original node representations is essential.
To address these objectives, we propose a balance loss $\mathcal{L}_{Bal}$ that optimizes the minimum transport distance between the distributions within each mini-batch, thereby constraining their similarity. We utilize the Wasserstein-1 distance $W\left(\mathbf{H^{r}},\mathbf{H^{g}}\right)$ from optimal transport theory to achieve this goal. Here, $\mathbf{H^{r}}$ and $\mathbf{H^{g}}$ denote two sets of representations, each comprising samples from a representation distribution.  The Wasserstein-1 distance is defined as follows: 

\begin{equation}
W\left(\mathbf{H^{r}},\mathbf{H^{g}}\right)=\inf _{\gamma \sim \Pi\left(\mathbf{H^{r}}, \mathbf{H^{g}}\right)} \mathbb{E}_{(x, y) \sim \gamma}[\|x-y\|],
\end{equation}
where $\Pi(\mathbf{H^{r}},\mathbf{H^{p}})$ represents the set of all joint distribution over $\mathbf{H^{r}}$ and $\mathbf{H^{p}}$, and $x, y$ represent the sample from $\mathbf{H^{r}}$ and $\mathbf{H^{p}}$ accordingly.

With the definition of Wasserstein-1 distance, the balance loss $\mathcal{L}_{Bal}$ can be defined as follows:
\begin{equation}
\mathcal{L}_{Bal}=W\left(\mathbf{H}, \mathbf{H^{+}}\right)+W\left(\mathbf{H}, \mathbf{H^{-}}\right).
\end{equation}
Since the exact Wasserstein-1 distance is difficult to calculate, we obtained an approximation of the Wasserstein-1 distance using the sinkhorn algorithm~\cite{cuturi2014fast}.




\noindent \textbf{Intra-graph contrastive loss.} For each subgraph $G_i$ sampled from the original graph, the subgraph representation is obtained through a readout function.
The readout function summarizes all the node representations $\{\mathbf{h}_j\}_{j\in \mathcal{I}}$ in a subgraph as the representation $\mathbf{s}_i$ of subgraph $G_i$. Here, we use the mean pooling function as the readout function and obtain the subgraph representation as:
\begin{equation}\label{eq:readout}
\mathbf{s}_{i} =\frac{1}{K} \sum_{j\in \mathcal{I}_i} \mathbf{h}_{j}.
\end{equation}

We treat the subgraph representation $\mathbf{s}_i$ for $G_i$ as a positive sample for node $i$ as in~\cite{jiao2020sub}. In addition, we shuffle the graph representation set including graph representation in Eq.~\eqref{eq:readout} for all subgraphs to generate negative examples.
More formally, we have
\begin{equation}
\left\{\widetilde{\mathbf{s}}_{1}, \widetilde{\mathbf{s}}_{2} \ldots, \widetilde{\mathbf{s}}_{N}\right\} \sim \mathbf{Shuffle}\left(\left\{\mathbf{s}_{1}, \mathbf{s}_{2}, \ldots, \mathbf{s}_{N}\right\}\right),
\end{equation}
where $\widetilde{\mathbf{s}}_{i}$ is the negative sample for node $i$.
We utilize the margin triple loss to push the representation of the central node towards that of the positive sample and pull it away from that of the negative example. The loss function for center node $i$ can be written as:
\begin{equation}
\mathcal{L}^i_{S} = \left[\sigma\left(\mathbf{h}_{i} \widetilde{\mathbf{s}}_{i}\right)- \sigma\left(\mathbf{h}_{i} \mathbf{s}_{i}\right)+\epsilon\right]_{+},
\end{equation}
where $\epsilon$ denotes margin value, $\sigma(\cdot)$ denotes the sigmoid function. Then, we can obtain the intra-graph contrastive loss $\mathcal{L}_{S}$ for the original graph as follows
\begin{equation}
\mathcal{L}_{S}=\frac{1}{N} \sum_{i=1}^{N}\mathcal{L}^i_{S}.
\end{equation}


The final loss function $\mathcal{L}_{Full}$ of the proposed subgraph contrastive learning framework can be written as:
\begin{equation}
\mathcal{L}_{Full}= \mathcal{L}_{S} + \alpha \mathcal{L}_{Bal} + \beta \mathcal{L}_{CL},
\end{equation}
where $\alpha$ is a hyperparameter to control the weight of the balance loss, $\beta$ is a hyperparameter to control the importance of the inter-graph contrastive learning loss $\mathcal{L}_{CL}$.

\subsection{Adversarial Curriculum Learning}
Curriculum learning \cite{bengio2009curriculum} is a training strategy that can improve the convergence rate and generalization ability of a machine learning model. The existing curriculum learning methods can hardly provide finer-grained and more efficient training for graph contrastive learning. In this subsection, we introduce our adversarial curriculum training framework which provides a coarse-grained sample partition based on pair-wise augmentations and a finer-grained training via an adversarial self-paced learning strategy.

Since the hyperparameter $\gamma$ in Eq. \eqref{equation-4} and Eq. \eqref{equation-5} ensures that the node pair search will only be performed when the sum of the distances between the two pairs of nodes is less than $2\gamma$, So, the value of $\gamma$ determines the number of node pairs to be used in pair-wise graph augmentation.
However, since the distribution of node pair distance, denoted as $\mathbf{Dis}$ might not be uniform, direct adjustment of $\gamma$ does not guarantee a linear increase in the number of node pairs to which the node pair search method is applied. If a good $\gamma$ increment is not chosen, the most extreme case will result in the distance between all node pairs is less than $2\gamma_{t}$ in epoch $t$ and greater than $2\gamma_{t+1}$ in epoch $t+1$. To solve this problem, we first estimate the distribution of distance $\mathbf{Dis}$ via analyzing node pairs in the set of subgraphs $\{G_i\}^N_{i=1}$. We define a \textit{quantile function} $q(\mathbf{Dis},\theta)$ that can calculate the $\theta$ quantile values from the distribution of distance $\mathbf{Dis}$. Then, we have:
\begin{equation}
\gamma=q(\mathbf{Dis},\theta).
\end{equation}
Through the quantile function $q$, we can control the value of the hyperparameter $\gamma$. Compared with manually setting $\gamma$, this method is data-driven and can automatically set a proper value for $\gamma$. It is suitable when we train models on a set of datasets with very different data distributions.

Based on this, we can use the parameter $\theta$ as a difficulty measurer to adjust the difficulty of the training data more stably. 
For training scheduling, we apply the widely used predefined pacing function in our framework \cite{hacohen2019power,penha2019curriculum}. To guarantee the quality of mirror graphs, we set the upper bound $M$ and lower bound $\theta_0$ for $\theta$. 
Then, we can modify the definition of the pacing function in \cite{hacohen2019power,penha2019curriculum} as follows:
\begin{equation}
\theta_{\text {linear}}(t)=\min \left(1,\frac{1}{M} \theta_{0}+\frac{1-\theta_{0}}{T} \cdot t\right)\cdot M.
\end{equation}
With the above pacing function, the parameter of $\theta$ would increase linearly as the training process.



However, the difficulty $\theta$ is coarse-grained because different subgraphs are sampled from different locations of the original graph, so even though these subgraphs are all in dataset $D_{\theta}=\left(G,G_{\theta}^{+},G_{\theta}^{-}\right)$ with $\theta$ of difficulty, they still have the difficulty of significant variance. Therefore, finer-grained graph curriculum learning should be sensitive to the difficulty of different subgraphs and further scheduling of samples of different difficulties for training. 
The self-paced learning strategy (SPL) \cite{kumar2010self} demonstrated in preliminaries is one way to solve this problem. 
However, SPL makes no further distinction for samples with loss below a threshold, and the same weights are applied to these samples. Some variants \cite{jiang2014easy,gong2018decomposition} of SPL assign higher weights to simpler samples that are below the threshold. However, we believe that this approach has a significant drawback, as training proceeds, the increment of difficult samples that are allowed to be trained is relatively small, but this increment of samples is critical to the performance of the model, so it is not appropriate to give these samples the same or even lower weights than other samples. Focusing too much on simpler samples can speed up the convergence but can also be a bottleneck of the model. To solve this problem, we propose a novel adversarial curriculum learning (ACL) strategy, which measures the difficulty via sample loss. In our ACL strategy, we define the node sample loss as follows:
\begin{equation}
\mathcal{L}^i=\mathcal{L}^i_{S} + \alpha \frac{1}{N} \mathcal{L}_{Bal}+\beta \mathcal{L}^i_{CL}.
\label{Loss}
\end{equation}
Different from the complete loss function $\mathcal{L}_{Full}$ assigning the same weight for each center node. In the ACL strategy, we assign different weights by optimizing the following objective function:
\begin{equation}
\max_{\boldsymbol{u} \in[0,1]^{N}}\min _{ \boldsymbol{v} \in[0,1]^{N}} \mathbb{E}(\boldsymbol{u} ,\boldsymbol{v} ; \boldsymbol{\lambda}) = \sum_{i=1}^{N} u_{i}v_{i} \mathcal{L}^i+\lambda_1 g(\boldsymbol{v})+\lambda_{2}f(\boldsymbol{u}),
\end{equation}
where $u_{i}v_{i}$ denotes the weights assigned to the $i$-th sample. $f$ and $g$ are the regularization terms on $\boldsymbol{u}$ and $\boldsymbol{v}$. $v_{i}$ ($u_{i}$) is used to control the importance of sample $i$ in training with large (small) difficulty. We put all the proofs of theorems and propositions into the appendix.

\begin{zltheo}
If $g(\boldsymbol{v})$ is convex on $[0,1]^N$ and $f(\boldsymbol{u})$ is concave on $[0,1]^N$, then there exists a unique optimal solution for the max-min optimization problem.
\end{zltheo}
This theorem guarantees the uniqueness of the optimal solution, which provides a high-level insight for designing the optimization algorithms. The alternating optimization algorithms can be an excellent method to solve the max-min optimization problem. In this paper, we adopt alternating optimization to update $\boldsymbol{u}$ and $\boldsymbol{v}$ one at a time. If we assume that the regularization for each node can be separated, the alternating optimization can be formulated as:
\begin{equation}\label{vstar}
v_{i}^{*}=\arg \min _{v_{i} \in[0,1]} u_{i}^{*} v_{i} \mathcal{L}^i+ \lambda_{1} g\left(v_{i}\right), \quad i=1,2, \cdots, N;
\end{equation}
\begin{equation}\label{ustar}
u_{i}^{*}=\arg \max _{u_{i} \in[0,1]} u_{i} v_{i}^{*} \mathcal{L}^i+\lambda_{2}f\left(u_{i}\right), \quad i=1,2, \cdots, N.
\end{equation}
With the optimal $v_{i}^{*},u_{i}^{*}$, we can rewrite the final loss as:
\begin{equation}
\mathcal{L}_{Full}=\sum_{i=1}^{N} u^*_{i}v^*_{i} \mathcal{L}^i.
\end{equation}

\begin{zlprop}
If $g(\boldsymbol{v})$ is convex on $[0,1]^N$ and $f(\boldsymbol{u})$ is concave on $[0,1]^N$, then the optimal solution $v_i^*$ decreases with $\mathcal{L}^i$ and $u_i^*$ increases with $v_i^*\mathcal{L}^i$. In addition, the optimal solution $u_i^*$ decreases with $\lambda_2$ and $v_i^*$ increases with $\lambda_1$.
\end{zlprop}
This proposition illustrates the characteristics of the optimal solution, which provides a deeper understating of the max-min optimization problem. The regularization functions $g$ and $f$ determine the strategy of re-weighting samples. In this paper, we propose two kinds of adversarial curriculum learning methods according to different regular functions, i.e., hard adversarial curriculum learning and soft adversarial curriculum learning.

\noindent \textbf{Hard Adversarial Curriculum Learning.} In hard adversarial curriculum learning (Hard ACL), we define both regular functions as $L_1$ regulation. The re-weighting strategy is to discard the samples with either too large difficulty or too small difficulty. More formally, we define the regular functions as follows,
\begin{equation}\label{eq:hard_label}
g(\boldsymbol{v} ; \lambda_{1})=-\lambda_{1} \sum_{i=1}^{N} v_{i}, \quad f(\boldsymbol{u} ; \lambda_{2})=-\lambda_{2} \sum_{i=1}^{N} u_{i}.
\end{equation}

Given any sample loss $\{\mathcal{L}^i\}^N_{i=1}$, we can obtain the optimal $u_{i}^{*}, v_{i}^{*}$ based on optimization in Eq. (\ref{ustar}) and Eq. (\ref{vstar}) in Appendix~\ref{the}.
\begin{zlprop}
If the regular functions are defined in Eq.~[\ref{eq:hard_label}]. Then, the optimal solution $u_{i}^{*}, v_{i}^{*}$ satisfy:
\begin{equation}
u_{i}^{*}v_{i}^{*}=\left\{\begin{array}{lr}
1, &  \lambda_{2}<\mathcal{L}^{i}<\lambda_{1}; \\
0, & \text {otherwise}.
\end{array}\right.
\label{hardacl}
\end{equation}
\end{zlprop}
According to the proposition, using $u^{*}v^{*}$ as sample weight allows the model to focus more on difficult samples close to the threshold and discard the overly easy samples, although of course the easy samples will be retrained if the model performance decreases.

\noindent \textbf{Soft adversarial curriculum learning.} By modifying the regularization function $f$ we can derive the soft adversarial curriculum learning (Soft ACL) strategy as follows:
\begin{equation}\label{eq:soft_label}
f(\boldsymbol{u} ; \lambda_{2})=-\frac{1}{2}\lambda_{2} \sum_{i=1}^{N} u_{i}^2.
\end{equation}


The optimal solution $u_{i}^{*}, v_{i}^{*}$ for soft adversarial curriculum learning satisfy the following condition:
\begin{equation}
u_{i}^{*}v_{i}^{*}=\left\{\begin{array}{lr}
1, & \lambda_{2}\leq \mathcal{L}^{i}<\lambda_{1};\\
\mathcal{L}^{i}/\lambda_{2}, &  \mathcal{L}^{i}<\lambda_{2}; \\
0, & \text {otherwise}.\\
\end{array}\right.
\label{softacl}
\end{equation}
If $u_{i}$ is fixed to $1$, ACL degenerates to the SPL. ACL allows us to discard overly easy samples. If training leads to an increase in the loss function for easy samples, the easy samples can also be included for the subsequent training epochs. Instead of discarding the overly easy samples, soft ACL assigns lower weights to easier samples. Thus, the training process would focus more on the difficult samples while having all the samples involved. Moreover, we summarize the process of the ACL strategy in Algorithm \ref{algorithm1} of Appendix~\ref{algorithm}.

%% file: samples/4Experiments.tex
\section{Experiment}
In this section, we conduct experiments on ACGCL for node classification using six different datasets. We then present the results of the ablation experiments and provide an analysis of pair-wise graph augmentation. We put the parameter analysis into the appendix.
\subsection{Experimental Settings}
\noindent \textbf{Datasets.}  We conducted experiments on six benchmark datasets, including Cora, Citeseer, PubMed \cite{kipf2016semi}, DBLP\cite{tang2008arnetminer}, Amazon-Computers \cite{shchur2018pitfalls} and Coauthor Physics. For the first three datasets, we use the standard data split provided by PyG. For the last three datasets, we follow the data split in \cite{zhu2020deep,zhu2022rosa}. The detailed statistics of these datasets are summarized in Table \ref{tab:datasets} of Appendix~\ref{Statistics}. 
 \begin{table*}[] 

	\renewcommand
 \arraystretch{1}
 \tabcolsep=0.3cm 
	\centering  
	\caption{Overall comparisons of node classification accuracy on six benchmark datasets}\label{table1}  
	\vspace{-3mm}
	\label{overall}  
	\begin{tabular}{p{56pt}ccccccc} 
		\toprule 
		Algorithm && Cora & Citeseer & PubMed & Amazon-Computer	 & DBLP & Coauthor-Phy\\
		\midrule 
		Raw features && 56.6 ± 0.4  &  57.8 ± 0.2 &  69.1 ± 0.2 & 73.8 ± 0.1 &71.6 ± 0.1&93.5 ± 0.0 \\  
		\hline
		GCN &&81.4 ± 0.6   & 70.3 ± 0.7 &  76.8 ± 0.6 &  84.5 ± 0.3&82.7 ± 0.2 & 95.6 ± 0.1 \\
		DGI && 82.3 ± 0.6  &  71.8 ± 0.7   & 76.8 ± 0.6 & 87.8 ± 0.2&83.2 ± 0.1 & 94.5 ± 0.5   \\
		GMI &&83.0 ± 0.3  & 73.0 ± 0.3 &  79.9 ± 0.2 &  82.2 ± 0.4&84.0 ± 0.1 & OOM \\
	    SUBG-CON && 82.1 ± 0.3 & 72.4 ± 0.4 & 79.2 ± 0.2 &89.1 ± 0.2&83.5 ± 0.2 & 95.5 ± 0.2\\
	    MVGRL&&82.9 ± 0.3 & 72.6 ± 0.4&80.1 ± 0.7& 87.5 ± 0.1& 69.2 ± 0.5 & 95.3 ± 0.0\\
	    GCA&& 81.8 ± 0.2 & 71.9 ± 0.4 & 81.0 ± 0.3 &88.9 ± 0.2&81.2 ± 0.1 & 95.6 ± 0.1  \\
        BGRL &&  82.7 ± 0.6 & 71.1 ± 0.8 & 79.6 ± 0.5 & 89.6 ± 0.3 & 83.7 ± 0.4 & 95.5  ± 0.1\\
        AFGRL &&  81.3 ± 0.2 & 68.7 ± 0.3 &  80.6 ± 0.4 & 89.5 ± 0.3 & 81.4 ± 0.2 & 95.7 ± 0.1\\
	    \hline
	    ACGCL (hard) && 83.7 ± 0.5 & 73.3 ± 0.4 & 80.9 ± 0.2 &89.5 ± 0.1 &83.9 ± 0.3 & 95.8 ± 0.0 \\
	    
	    ACGCL (soft) && $\boldsymbol{84.4}$  ± $ \boldsymbol{0.6^*}$  & $ \boldsymbol{73.5}$ ± $ \boldsymbol{0.3^*}$  &  $ \boldsymbol{81.4}$ ± $ \boldsymbol{0.2^*}$ & $ \boldsymbol{89.7}$ ± $ \boldsymbol{0.2^*}$ &$ \boldsymbol{84.2}$  ± $\boldsymbol{0.2^*}$ & $\boldsymbol{95.9^*}$ ± $\boldsymbol{0.1}$  \\
		\bottomrule 
	\end{tabular}
	\\``\textbf{{\Large *}}'' indicates the statistically significant improvements (i.e., two-sided t-test with $p<0.05$) over the best baseline.
\vspace{-3.3mm}
\end{table*}

\noindent \textbf{Baselines.} We compare our method with 2 node-graph GCL methods:
DGI \cite{velickovic2019deep}, SUBG-CON \cite{jiao2020sub}, and 3 node-node GCL methods: GMI \cite{peng2020graph}, GCA \cite{zhu2021graph}, MVGRL \cite{hassani2020contrastive}, BGRL~\cite{thakoor2021large}, AFGRL~\cite{lee2022augmentation}.
 
\noindent \textbf{Implementation Details.} ACGCL uses 2-layer GCN~\cite{kipf2016semi} as encoder for the Citesser and DBLP datasets, and 1-layer GCN for the other datasets. For all datasets, the dimension of node representation is set to $1,024$. For subgraph sampling, we set the subgraph size to $20$ by default. During training, $500$ subgraphs are randomly sampled in each mini-batch. We use Adam optimizer with an initial learning rate of $0.001$. The starting difficulty $\theta_{0}$ of curriculum learning is set to $15$ and the maximum difficulty $M$ is set to $50$. Instead of manually setting thresholds, we set $\lambda_{1}$ and $\lambda_{2}$ to the median and $0.95$ times the minimum of the sample loss for the first step. This makes the model easier to migrate to different datasets. We avoid duplicating in processing node pairs in pair-wise graph augmentation. We let SUBG-CON share the same encoder and subgraph size as ACGCL for a fair comparison.
%


\noindent \textbf{Evaluation Metrics.} We evaluate the performance of our proposed framework on one of the most widely used downstream tasks -- node classification. Following previous works \cite{jiao2020sub,mo2022simple,zhu2022rosa}, we train a simple logistic regression classifier utilizing the node representation learned by the ACGCL framework and the baselines. We set the metrics of performance to be the node classification accuracy. 

\subsection{Overall Performance}
We compare the node classification accuracy performance of different methods on five datasets in Table \ref{table1}. We can observe that:
\begin{itemize}[leftmargin=*]
\item ACGCL (soft) achieves the best performance on all datasets. Different from those methods (e.g., SUBG-CON, GCA, and MVGRL) that utilize data augmentation by subgraph sampling or predefined graph augmentation (e.g., node dropping and edge perturbation), ACGCL generates more controllable positive and negative samples by pair-wise graph augmentation, which are shown to result in better performance in graph contrastive learning.
\item ACGCL (hard) has slightly lower accuracy compared with ACGCL (soft), but it still outperforms most of the baselines. The performance of ACGCL (hard) would be extremely limited when the difficulty of samples in a dataset does not vary too much (e.g., PubMed). It is difficult for ACGCL (hard) to set appropriate thresholds to filter out easy samples, resulting in low sample efficiency. 
\item ACGCL (soft) achieves a significant performance improvement on the Cora dataset. The reason is that the Cora dataset has low overall sample difficulty. The baseline methods learn from all samples with equal weights and suffer from the serious issue of the sparsity of difficult samples. 
\end{itemize}
\begin{table*}[] 
	\renewcommand\arraystretch{1}\tabcolsep=0.3cm
	\caption{Results of ablation study}  
	\label{Table2}  
	\vspace{-3mm}
	\begin{tabular}{p{70pt}ccccccc} 
		\toprule 
		 && Cora & Citeseer & PubMed & Amazon-Computer & DBLP & Coauthor-Phy \\
		\midrule 
		Non Curr&& 83.3 ± 0.3 & 73.0 ± 0.3& 79.9 ± 0.2 & 88.9 ± 0.3 &83.3 ± 0.1 & 95.5 ± 0.1 \\
		Curr&& 83.5 ± 0.2 &72.9 ± 0.6 & 81.0 ± 0.2 & 89.2 ± 0.2 & 83.6 ± 0.3 & 95.8 ± 0.1 \\
		Curr SPL && 83.0 ± 0.3   & 72.8 ± 0.3 & 80.3 ± 0.3 & 88.5 ± 0.2 &83.4 ± 0.2 & 95.6 ± 0.1  \\
        \midrule
		Soft ACL w/o $\mathcal{L}_{Bal}$&& 84.2 ± 0.4& 72.7 ± 0.3 & 80.8 ± 0.3 & 89.6 ± 0.1 & 83.9 ± 0.3  &  95.8 ± 0.1 \\
        Soft ACL w/o $\mathcal{L}_{CL}$&& 82.8 ± 0.3 & 72.6 ± 0.6 & 79.8 ± 0.4 &89.3 ± 0.2 & 83.7 ± 0.3 &95.8 ± 0.0 \\
        \midrule
		Soft ACL && $\boldsymbol{84.4}$ ± $\boldsymbol{0.6}$  &  $\boldsymbol{73.5}$ ± $\boldsymbol{0.3}$   &$\boldsymbol{81.4}$ ± $\boldsymbol{0.2}$ &$ \boldsymbol{89.7}$ ± $ \boldsymbol{0.2}$&  $\boldsymbol{84.2}$ ± $\boldsymbol{0.2}$ & $\boldsymbol{95.9}$ ± $\boldsymbol{0.1}$ \\ 
		\bottomrule 
	\end{tabular}
	\vspace{-2.6mm}
\end{table*}
\subsection{Ablation Study}
In this subsection, we perform ablation study, where the components of the proposed method are removed one by one to verify the effectiveness of each of them. 
We report results on all datasets.
Soft ACL stands for the soft version of adversarial curriculum learning. Curr SPL denotes the model where we remove the adversarial curriculum learning process. In particular, Curr SPL only uses the original curriculum learning to control the pair-wise data augmentation and self-paced learning to adjust the proportion of samples used in training. 
Curr refers to the model that only uses the original curriculum learning to control pair-wise data augmentation without self-paced learning.
Non Curr is the model that uses pair-wise data augmentation without curriculum learning or self-paced learning. Soft ACL w/o $\mathcal{L}_{Bal}$ and Soft ACL w/o $\mathcal{L}_{CL}$ denotes the model without using balance loss and contrastive learning loss. Note that the results of Hard ACL have been given in Table \ref{overall}. From Table \ref{Table2}, we can find that the performance improvement of the ACL strategy is obvious. It is worth noting that Curr SPL has the worst performance, which further verifies that the model that focuses excessively on simple samples might be harmful to its performance. We can also observe that $\mathcal{L}_{Bal}$ and $\mathcal{L}_{CL}$ are crucial to the performance of our model. The parameter analysis can be found in Appendix~\ref{ana1}, We investigate the performance of the model when $\alpha$, $\beta$ in Eq.~(\ref{Loss}), and subgraph size $K$ are varied.

\subsection{Pair-wise Graph Augmentation}
In this subsection, we do some empirical analysis of pair-wise graph augmentation, including graph difficulty validation, distance function selection, and subgraph embedding distribution analysis. The details on the latter two can be found in Appendix~\ref{Analysis}.

\begin{figure}[]
	\centering
	{\subfigure{\includegraphics[width=0.409\linewidth]{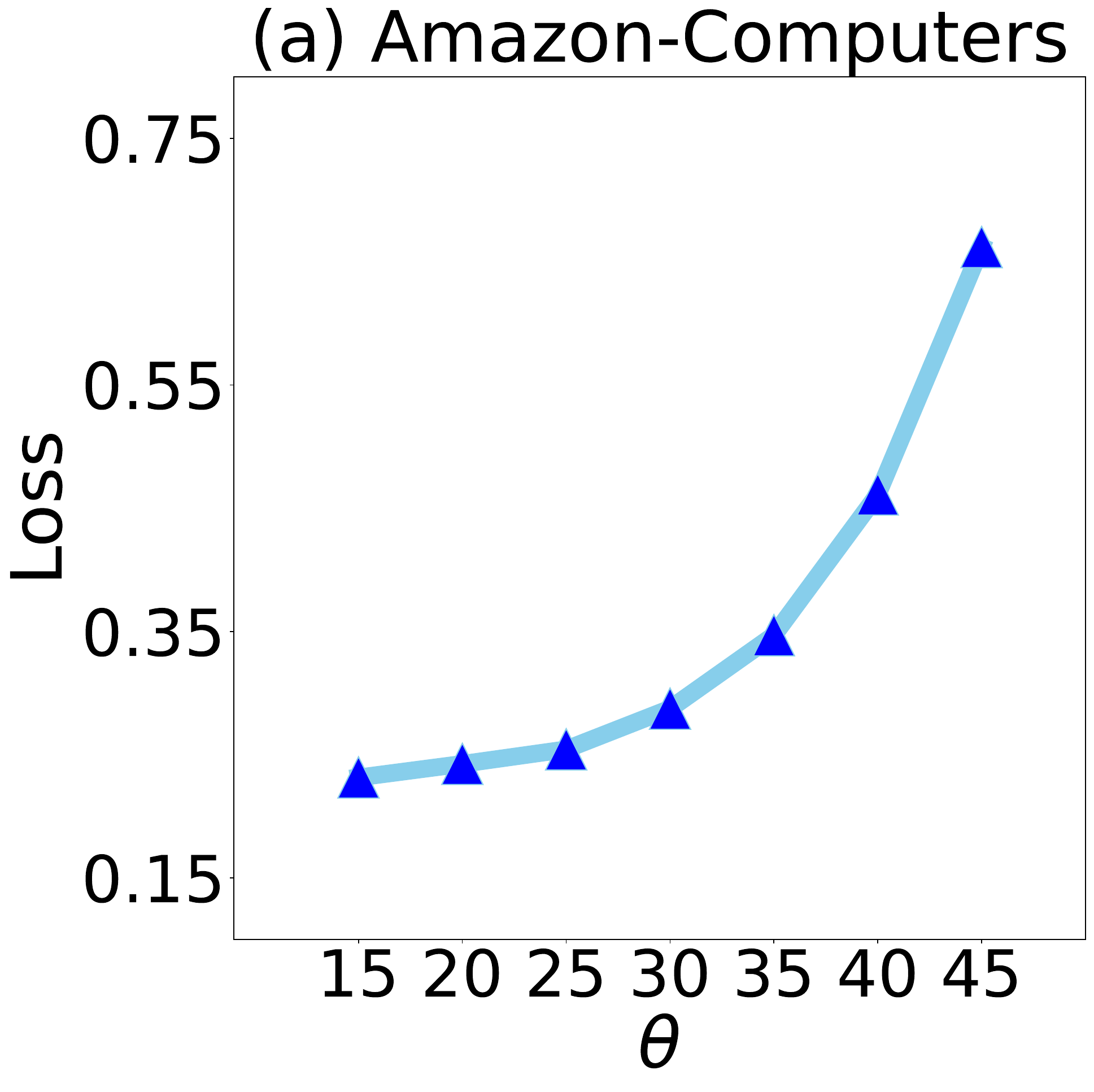}}}
	{\subfigure{\includegraphics[width=0.409\linewidth]{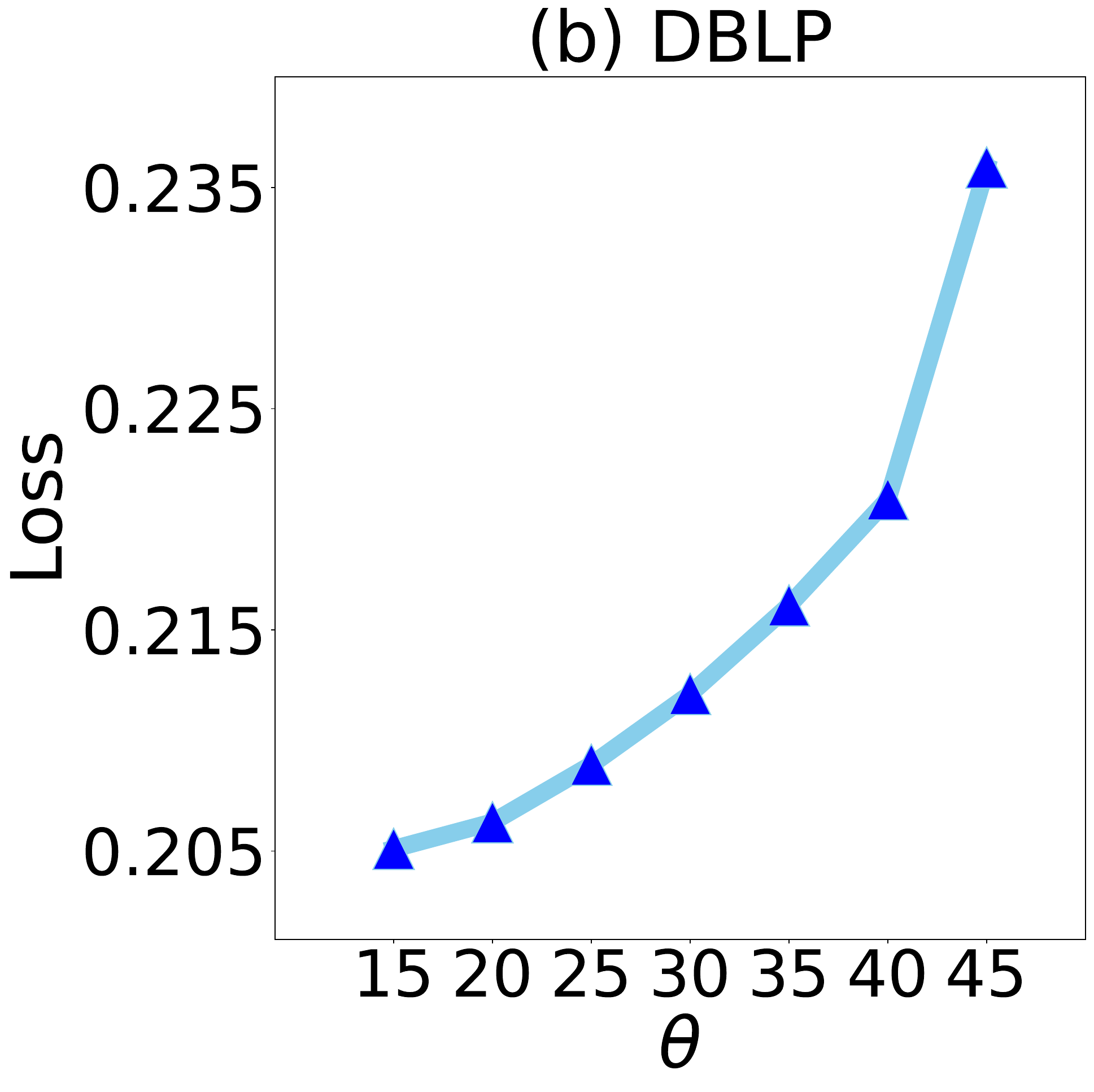}}}
	\vspace{-5mm}
	\caption{Graph sample difficulty validation results. With larger $\theta$, the sample difficulty of mirror subgraphs from pair-wise graph augmentation increases. }\label{fig:training}\label{fig4}
\vspace{-3mm}
\end{figure}

\noindent \textbf{Graph sample difficulty validation.} 
%
To investigate the relationship between the difficulty of samples from pair-wise graph augmentation and the parameter $\theta$, we utilize GNN models pre-trained on Amazon Computers and DBLP datasets. We compute representations of positive and negative mirror graphs generated with varying $\theta$ values, as well as the original graph. The triple loss is used as an evaluation metric for sample difficulty. The results, depicted in Figure \ref{fig4}, indicate that as $\theta$ increases, the samples generated by pair-wise graph augmentation become more difficult.

%% file: samples/2Relatedwork.tex
\section{RELATED WORK}
\noindent\textbf{Graph Contrastive Learning.} GraphCL \cite{you2020graph} proposes four simple data enhancement methods to obtain positive samples, including node dropping, edge perturbation, attribute masking, and subgraph sampling, while GraphCL uses random negative sampling to construct negative sample graphs.
Based on GraphCL, a series of graph contrast learning methods are proposed.
%
%
To extend data augmentation methods based on predefined perturbations (e.g., those proposed by GraphCL), there are works that aim to generate different views through generative graph machine learning models such as GVAE~\cite{you2022bringing,simonovsky2018graphvae}. 
However, most previous work ignores the fact that the randomly sampled negative samples can be false negatives since they may come from similar data distributions.
Recently, some work starts to work on negative sample selection.
DEC~\cite{chuang2020debiased} performs clustering along with the training process to generate pseudo node labels. Then negative samples are selected based on their pseudo labels. 
PGCL~\cite{lin2022prototypical} also adopts the idea of generating pseudo labels by clustering. The clustering is carried out on the representations of subgraphs. It calculates the weight of different samples such that the training processes focus on the samples in clusters that are moderately distance from the center of the positive sample clusters. This is because the samples with too large (too small) distances are too easy (too difficult), which can be ineffective for representation learning through graph contrastive learning. 
However, these methods still rely on predefined graph augmentation methods. Different from them, this work proposes a novel pair-wise graph augmentation approach that enables controlling the similarity of positive samples and the difficulty of negative samples. There are also works~\cite{lin2022spectral,liu2022revisiting} that try to explain the effectiveness of graph contrastive learning from the spectral perspective, and they are orthogonal to our branch of research.

\noindent\textbf{Curriculum Learning.} Curriculum learning \cite{bengio2009curriculum} aims to imitate human learning, where the model starts with learning from easy samples and gradually learns to handle harder samples as the model becomes more and more capable.
Curriculum learning methods consist of two major components: a difficulty measurer and a training scheduler \cite{wang2021survey,soviany2022curriculum}. The difficulty measurer is the foundation of curriculum learning. It defines the difficulty of each sample so that the training scheduler can add samples to the training process in ascending order of difficulty.
There are two families of difficulty measurers: the predefined approaches (e.g., Mahalanobis distance~\cite{el2020student}) and the data-driven approaches (e.g., scoring models). 
PinSage \cite{ying2018graph} adopts curriculum learning for graph representation learning. It uses the similarity between the query item and negative samples as the difficulty measure. In particular, PinSage computes personalized Pagerank (PPR) to rank other items w.r.t. the query item, and randomly selects difficult negative samples at the tail of the ranking. CuCo \cite{chu2021cuco} uses cosine similarity of graph representations as the difficulty measurer to calculate similarity scores of graphs, and obtains a sequence of samples from easy to hard based on the similarity scores. 
However, Pinsage measures node-level difficulty, it does not have the ability to measure graph-level difficulty. CuCo can calculate graph-level difficulty, but the value of difficulty depends on a single positive sample, so it cannot measure difficulty strictly.
The design of the training scheduler is mainly divided into predefined pacing function \cite{hacohen2019power,penha2019curriculum} and automatic approaches \cite{kumar2010self,florensa2017reverse}. The pacing function is an incremental function, which outputs the proportion of training samples allowed. The automatic method is represented by self-paced learning~\cite{kumar2010self}, which adjusts the proportion of training samples in real-time by training loss. However, self-paced learning and its variants~\cite{jiang2014easy,zhao2015self} focus on simple samples excessively, which results in the sparsity problem of difficult samples and impedes the learning process. 
Different from the above methods, our ACGCL framework can enhance the focus on a small number of difficult samples via reweighting the loss while gradually increasing the overall difficulty of training data.

%% file: samples/5Conclusion.tex
\section{Conclusion}
In this paper, we proposed a novel Adversarial Curriculum Graph Contrastive Learning framework, called ACGCL, which consists of three modules. The first is pair-wise graph augmentation, which perturbs graph structures to construct controllable positive and negative mirror graphs. The second is the subgraph contrastive learning, which consists of three types of loss for subgraphs. It reduces the complexity of learning graph representations for large graphs. The third is the novel adversarial curriculum training, which assigns more weight to difficult samples via adversarial learning in order to alleviate the sparsity problem of difficult samples in traditional curriculum learning. We verify the performance of our ACGCL via extensive experiments on six benchmark datasets.  We would like to explore more flexible pair-wise graph augmentation such as model-based graph augmentation strategies in the future.  